\documentclass[a4paper,conference]{IEEEtran}
\IEEEoverridecommandlockouts

\usepackage[style=ieee,backend=biber]{biblatex}
\usepackage{amsmath,amssymb,amsfonts}
\usepackage{graphicx}
\usepackage{booktabs}
\usepackage{url}
\usepackage[hidelinks]{hyperref}
\usepackage{mathtools}
\usepackage{subcaption}
\usepackage{standalone} 
\usepackage[dvipsnames]{xcolor}
\usepackage{graphicx} 
\usepackage{amsmath}
\usepackage{amssymb}
\usepackage{amsfonts}
\usepackage{dsfont}
\usepackage{tikz}
\usepackage{pgfplots}
\pgfplotsset{compat=1.18}
\usepgfplotslibrary{fillbetween}
\usepackage{threeparttable}
\usepackage{enumitem}
\usepackage{algorithm}
\usepackage{algpseudocode}
\usepackage{booktabs, multirow,xcolor,colortbl, makecell}
\usepackage{multirow}

\pgfplotsset{
    colormap={reds}{
        rgb255(0cm)=(255,255,255);
        rgb255(1cm)=(254,224,210);
        rgb255(2cm)=(252,146,114);
        rgb255(3cm)=(222,45,38)
    }
}
\pgfplotsset{
    colormap={redyellowgreen}{
        rgb255(0cm)=(220,20,60);   
        rgb255(1cm)=(255,255,160); 
        rgb255(2cm)=(34,139,34)    
    }
}

\newcommand{\authorfont}{\fontsize{11pt}{10pt}\selectfont}
\renewcommand\IEEEkeywordsname{Keywords}

\begin{document}

\title{Transformer-based Multi-agent Reinforcement Learning for Separation Assurance in Structured and Unstructured Airspaces}

\author{\IEEEauthorblockN{\authorfont Arsyi Aziz}
\IEEEauthorblockA{\normalsize Department of Computer Science \\
\normalsize George Washington University \\
\normalsize Washington D.C., United States \\
}
\and
\IEEEauthorblockN{\authorfont Peng Wei}
\IEEEauthorblockA{\normalsize Department of Mechanical and Aerospace Engineering \\
\normalsize George Washington University \\
\normalsize Washington D.C., United States\\
} 
}

\maketitle

\begin{abstract}
Conventional optimization-based metering depends on strict adherence to precomputed schedules, which limits the flexibility required for the stochastic operations of Advanced Air Mobility (AAM). In contrast, multi-agent reinforcement learning (MARL) offers a decentralized, adaptive framework that can better handle uncertainty, required for safe aircraft separation assurance. Despite this advantage, current MARL approaches often overfit to specific airspace structures, limiting their adaptability to new configurations. To improve generalization, we recast the MARL problem in a relative polar state space and train a transformer encoder model across diverse traffic patterns and intersection angles. The learned model provides speed advisories to resolve conflicts while maintaining aircraft near their desired cruising speeds. In our experiments, we evaluated encoder depths of 1, 2, and 3 layers in both structured and unstructured airspaces, and found that a single encoder configuration outperformed deeper variants, yielding near-zero near mid-air collision rates and shorter loss-of-separation infringements than the deeper configurations. Additionally, we showed that the same configuration outperforms a baseline model designed purely with attention. Together, our results suggest that the newly formulated state representation, novel design of neural network architecture, and proposed training strategy provide an adaptable and scalable decentralized solution for aircraft separation assurance in both structured and unstructured airspaces.\end{abstract}

\let\thefootnote\relax
\footnotetext{%
  \noindent%
  \rule{0.98\linewidth}{0.1pt}\par
  A demonstration of the single-layer transformer model performance is available at
  \url{https://youtu.be/DpwM6QDnCuU}.%
}

\begin{IEEEkeywords}
Advanced Air Mobility, Separation Assurance, Multi-agent Reinforcement Learning, Transformers
\end{IEEEkeywords}

\section{Introduction}

Contemporary air traffic management (ATM) systems are built around optimization-based schedules. For instance, in the United States, Traffic Management Advisor and Time-Based Flow Management (TMA/TBFM) use schedules to sequence aircraft at metering fixes. While these scheduling approaches work well for current air traffic operations, particularly for maintaining safe aircraft separation, they may not be suitable for the emerging demands of Advanced Air Mobility (AAM). Point-to-point flights, higher traffic volumes, and the rapidly changing, stochastic airspace conditions may invalidate computed schedules, thereby reducing their capacity to address future air traffic conditions.

Learning-based methods, particularly reinforcement learning (RL), offer a promising alternative. Compared to schedule-based approaches, RL can learn policies that react in real time to dynamic and uncertain air traffic. In single-encounter settings, it has been shown to learn effective vectoring strategies for conflict resolution \cite{Pham2019Machine}. At a broader control scale, RL can be extended to multi-agent reinforcement learning (MARL), to also support cooperative, en-route separation assurance among multiple aircraft in both structured \cite{brittain2022marl} and unstructured airspaces \cite{groot2025comparing}.

Despite recent advances, generalization capability remains a key barrier to MARL in this domain. Models trained on a narrow set of scenarios may often overfit to specific airspace-structure parameters, which consequently reduces their robustness and efficacy as the airspace structure or traffic density changes. A MARL policy trained for one airspace structure may fail in another, and one optimized for moderate traffic may become unsafe under high density or overly conservative when traffic is light. This susceptibility to scenario or distribution shift underscores the need for methods that better generalize across airspace configurations and traffic densities.

To address this limitation, we focus on the aircraft separation assurance problem and introduce a MARL framework that improves generalization across both structured and unstructured airspaces for AAM operations. Building on the encoding capacity of transformers and the adaptability of MARL, our method encodes both ownship and intruder states as tokens within a transformer encoder, which then outputs speed advisories in the form of acceleration or deceleration recommendations. Instead of training aircraft (or agents) solely on a single route structure or fixed traffic pattern, we design a diverse training procedure that exposes a learning agent to different airspace structures and traffic densities. By exposing agents to this variety, they learn representations that transfer effectively to new airspace structures unseen during training.

Additionally, to further enhance generalization, we modify the training pipeline with changes to both the reinforcement learning problem formulation and the neural network architecture. For the problem formulation, we define a state space in relative polar coordinates to emphasize ego-centric states, and we augment the reward with a speed-incentive term. For the neural network architecture, we shift from attention to tokenization by introducing a classifier token conditioned on the ownship information. 

Therefore, the main contributions of this work are:
\begin{enumerate}
\item We reformulate the reinforcement learning problem using a polar state representation and a speed-incentivized reward function;
\item We propose an encoder–transformer architecture with a conditioned classifier token that scales to a variable number of intruder aircraft; and
\item We design a training regime based on procedurally generated sector structures with varying intersection angles and traffic densities.
\end{enumerate}
Together, these contributions support the generalization of separation assurance models for future AAM operations.

The remainder of this paper is organized as follows. \textit{Section~\ref{sec:related-work}} reviews related work. \textit{Section~\ref{sec:problem-formulation}} formalizes the separation assurance problem as a Markov decision process. \textit{Section~\ref{sec:network-structure} }introduces our modified transformer-encoder architecture. \textit{Section~\ref{sec:training}} presents the model training details. \textit{Section~\ref{sec:results-discussions}} reports the results of the experiments and \textit{Section~\ref{sec:discussions}} provides an accompanying discussion. \textit{Section~\ref{sec:conclusion}} concludes with a discussion of the broader implications of our work and proposes avenues for future research.

\section{Related Work}\label{sec:related-work}
\subsection{Transformer Networks}
Transformers~\cite{vaswani2017attention} are sequential models that operate on sequences of discrete tokens. For example, in natural language processing, tokens may correspond to words or sub-words, while in aircraft separation assurance and conflict resolution, a token represents an encoding of an aircraft's state vector. The core of the transformer architecture is the self-attention mechanism, which allows a model to compare and combine information from all token positions. This stands in contrast to recurrent architectures such as LSTMs~\cite{hochreiter1997long} and GRUs~\cite{cho-2014-learning}, which process tokens sequentially and often struggle either to preserve information from early positions or to parallelize computation over the sequence~\cite{vaswani2017attention}. Encoder-only architectures such as BERT~\cite{devlin2019bert} build on the transformer architecture to map an input sequence to contextualized, higher-level token embeddings. These token-level representations can then be aggregated in different ways for downstream classification tasks~\cite{devlin2019bert, dosovitskiy2020image}, for example, by introducing a dedicated classification token~\cite{devlin2019bert} or by applying pooling operations such as mean or max pooling over tokens~\cite{reimers2019sentence}.

\subsection{Multi-agent Reinforcement Learning for Aircraft Separation Assurance}
Multi-agent reinforcement learning (MARL) is an approach to machine learning in which a model is trained to produce desired behaviors by interacting with an environment under a prescribed reward function. In the separation assurance problem, the environment is modeled as multiple aircraft interacting within a shared airspace. Early MARL-based approaches in this setting represented the policy with a fully connected network and used input padding to handle a variable number of aircraft~\cite{brittain2019marl}. This design imposes a fixed upper bound on the number of aircraft that the model can process. For example, if the input layer is configured to process features for up to ten aircraft, the model cannot accommodate scenarios with more than the prescribed ten aircraft.

Subsequent work introduced attention mechanisms to better handle varying traffic levels~\cite{brittain2022marl}. These models learn the relative importance of each intruding aircraft, which allows them to accommodate a changing number of aircraft. This approach was later extended with transformer architectures and further improved by encoding each intruder’s state relative to the ownship~\cite{groot2022using}. Building on these developments, our work proposes modifications that further enhance adaptability to different airspace configurations and traffic densities.
\section{Problem Formulation}\label{sec:problem-formulation}
We study a separation assurance problem for AAM operations in a low-altitude sector, with extensions to unstructured airspace. In this problem, we consider the scenarios in which multiple aircraft attempt to maintain different desired cruising speeds. To prevent collisions, a separation assurance model must monitor the airspace for potential conflicts and issue discrete speed advisories (reduce, hold, increase) to maintain safe separation. Because desired cruising speeds are also taken into account, when safety permits, the model should ensure that each aircraft maintains proximity to these speeds. The problem is therefore posed in a speed-controlled separation assurance setting with two goals: \textit{(i)} maintain safe separation of aircraft and \textit{(ii)} minimize deviation from each flight's desired cruising speed.

\subsection{Safety-Critical Events}
We define three categories of safety-critical events that may arise within an airspace:

\paragraph{Conflict}
A conflict is a predicted breach of separation minima within a specified look-ahead horizon. In other words, if aircraft maintain their current trajectories and speeds, they are expected to lose safe separation within the defined time window. In our speed-only control setting, the agent must issue acceleration or deceleration commands to mitigate these predicted infringements proactively.
\paragraph{Loss of Separation}
A loss of separation (LoS) occurs when the separation minima are actually violated, meaning two or more aircraft simultaneously occupy the protected zone. Upon detection of a LoS event, the aircraft must immediately coordinate speed adjustments (either an increase or decrease) to restore separation and reduce the risk of escalation toward a near mid-air collision (NMAC).

\paragraph{Near Mid-Air Collision}
A near mid-air collision (NMAC) is the most severe event, defined as two or more aircraft coming within an unsafe distance of one another. The occurrence of an NMAC indicates a failure of the air traffic control system to maintain safety within the airspace. If speed advisories cannot fully resolve the conflict within the critical thresholds, the onboard collision-avoidance system---Traffic
Alert and Collision Avoidance System (TCAS) or Airborne Collision Avoidance System X (ACAS-X)---will be activated to issue coordinated vertical resolution advisories (e.g., climb or descend) to avoid collisions. 

\subsection{Mathematical Formulation}
We model sector–level coordination of multiple aircraft (agents) as a fully observable, multi-agent Markov decision process with a time-varying population. The process is described by the tuple $$
\mathcal{M} \coloneq \langle \mathcal{N}_{\max}, \mathcal{S}, \{\mathcal{A}_i\}_{i\in\mathcal{N}_{\max}}, \mathcal{P}, \mathcal{R}, \gamma\rangle,
$$
where $\mathcal{N}_{\max}$ indexes a superset of possible aircraft, $\mathcal{S}$ is the state space, $\mathcal{A}_i$ is the action space of agent $i$, $\mathcal{P}$ is the transition kernel (determined by aircraft dynamics and atmospheric conditions), $\mathcal{R}$ is the reward function, and $\gamma\in[0,1]$ is the discount factor. Furthermore, let $\mathcal{N}_t\subseteq\mathcal{N}_{\max}$ denote the set of active aircraft at time $t$.

At each discrete time $t$, the environment has a global state $s_t\!\in\!\mathcal{S}$. From this global state, each active aircraft $i\in\mathcal{N}_t$ observes an ego-relative state
$$
s_t^{(i)} \coloneq \psi_i(s_t)\in\mathcal{S}_i,
$$
where $\psi_i:\mathcal{S}\!\to\!\mathcal{S}_i \subset S$ is a deterministic, agent-specific projection that extracts the states of aircraft $i$ from the global state expressed in $ i$'s relative frame. ADS–B and error-free communications assume these states are accurate.

Each aircraft then selects an action according to a policy
$$
a_t^{(i)} \sim \pi_i(\cdot \mid s_t^{(i)}),
$$
and we write the joint action profile as the set of all actions
$
a_t \coloneq \big\{a_t^{(i)}\big\}_{i\in\mathcal{N}_t}.
$
The next state evolves as
$$
{s_{t+1} \sim \mathcal{P}(\cdot \mid s_t, a_t),
}$$
which implicitly allows for arrivals/departures of aircraft, so that the number of aircraft may change and $\mathcal{N}_{t+1}$ need not equal $\mathcal{N}_t$ in general.

Following each transition, aircraft $i$ receives a reward
$$
r_t^{(i)} \coloneq \mathcal{R}_i\!\big(s_t^{(i)}, a_t^{(i)}\big). 
$$
We denote $r_t \coloneq \{r_t^{(i)}\}_{i\in\mathcal{N}_t}$ as the collection of all rewards obtained at timestep $t$. A trajectory is denoted as
$
{\tau = (s_0, a_0, r_0, s _1, a_1, r_0\dots, s_T),}
$
defined over a finite horizon $T$. Completing a trajectory, for agent $i$, the expected discounted return is
$$J_i(\pi) \coloneq \mathbb{E}_{\tau\sim\pi}\!\left[\sum_{t=0}^{T-1}\gamma^t\, r_t^{(i)}\right].$$ 
Since we follow a general-sum multi-agent setting, we maximize the average of these values across all aircraft, so the overall objective is to find the maximizing policy
$$\pi^\star = \arg\max_{\pi} \frac{1}{|\mathcal{N}_\text{max}|}\sum_{i\in\mathcal N_\text{max}} J_i(\pi).$$

\subsection{State Space} \label{sec:state-space}
We construct the agent’s state space in an egocentric frame, consisting of both ownship features and a set of intruder features.

\begin{enumerate}
    \item \textit{Ownship Features}\\
    The ownship features provide the agent with speed-related information, consisting of the following elements:
    \begin{itemize}
        \item \textit{Calibrated airspeed}: the airspeed of aircraft $i$ at time $t$, corrected for instrument and positional errors, denoted by $v_{\text{cas}, t}^{(i)}$. 
        
      \item \textit{Speed deviation}: the absolute difference between the calibrated airspeed and the desired speed of aircraft $i$ at time $t$, defined as
        \begin{equation*}
        \Delta v_{t}^{(i)} \coloneq \left\lvert v_{\text{cas}, t}^{(i)} - v_{\text{des}}^{(i)} \right\rvert.
        \end{equation*}
    \end{itemize}

    
    \item \textit{Intruder Features}\\
    The intruder features are the set of intruder information encoded relative to the ownship, which includes:

    \begin{itemize}
    \item \textit{Relative position}: the position of each intruder relative to the ownship, expressed in polar coordinates. Let $d^{(i,j)}_t$ denote the Euclidean distance between aircraft $i$ and $j$ at time $t$. The relative position encompasses four components:
    \begin{itemize}
        \item \textit{Distance to the near mid-air collision (NMAC) boundary}:
        \begin{equation*}
            d_{\text{nmac}, t}^{(i,j)} \coloneq  d_t^{(i,j)} - r_\text{nmac},
        \end{equation*}
        where $r_\text{nmac}$ denotes the NMAC radius of $500$ feet.
        
        \item \textit{Distance to loss of separation (LOS) boundary}:
        \begin{equation*}
            d_\text{pz, t}^{(i,j)} \coloneq d_t^{(i,j)} - r^{(i)}_\text{rpz},
        \end{equation*}
        where $r_\text{pz}^{(i)}$ is the protection zone radius for loss of separation of aircraft $i$, which is by default set to $5$ nautical miles. 
        
       \item \textit{Relative bearing angle of intruder $j$ with respect to ownship $i$:}
        \begin{multline*}
        \theta^{(i,j)}_t \coloneq
        \big(\texttt{arctan2}(\Delta y_t^{(i,j)}, \Delta x_t^{(i,j)}) - \phi^{(i)}_t\big).
        \end{multline*}
        Here, $\Delta x_{t}^{(i,j)}$ and $\Delta y_{t}^{(i,j)}$ are the relative position components of intruder $j$ with respect to ownship $i$. The function $\texttt{arctan2}$ is the four-quadrant inverse tangent, which returns the absolute bearing of the intruder to the ownship. Subtracting the ownship heading $\phi^{(i)}_t$ yields the relative bearing. This feature is encoded using the sine and cosine functions to eliminate angular discontinuities in the features.

        \item \textit{Loss-of-separation indicator}:
        \begin{equation*}
            b_\text{los,t}^{(i)} \coloneq \mathds{1}_{\{ d^{(i,j)}_t \leq r_\text{pz} \}},
        \end{equation*}
        a binary variable that equals $1$ if aircraft $i$ and $j$ are within the protection zone and $0$ otherwise. Although this condition can be inferred from $d_t^{(i,j)}$ relative to $r^{(i)}_\text{pz}$,  the explicit inclusion of the indicator serves two purposes: \textit{(i)} it provides a direct encoding of loss-of-separation events, and \textit{(ii)} it allows the model to selectively deactivate or modulate other features when such loss-of-separation events occur.
        
        \end{itemize}
        
        \item \textit{Relative velocity}. In addition to positional features, we also include two velocity features, namely radial velocity and tangential velocity. To define these features, at time $t$, denote the relative position and velocity between ownship $i$ and intruder $j$ as
$$
\mathbf p_t^{(i,j)} \coloneq
\begin{bmatrix}
\Delta x_t^{(i,j)} \\
\Delta y_t^{(i,j)}
\end{bmatrix},
\qquad
\mathbf v_t^{(i,j)} \coloneq
\begin{bmatrix}
\Delta v_{x,t}^{(i,j)} \\
\Delta v_{y,t}^{(i,j)}
\end{bmatrix},
$$
where $\Delta x_t^{(i,j)} = x_t^{(j)}-x_t^{(i)}$ and $\Delta y_t^{(i,j)} = y_t^{(j)}-y_t^{(i)}$ are the east–west and north–south position differences, and $\Delta v_{x,t}^{(i,j)} = v_{x,t}^{(j)}-v_{x,t}^{(i)}$, $\Delta v_{y,t}^{(i,j)} = v_{y,t}^{(j)}-v_{y,t}^{(i)}$ are the corresponding velocity-component differences.

The radial unit vector, pointing from $i$ to $j$, is
$$
\hat{\mathbf e}_{p,t}^{(i,j)} \;\coloneq\; \frac{\mathbf p_t^{(i,j)}}{\lVert \mathbf p_t^{(i,j)} \rVert},
$$
and the tangential unit vector is obtained by a counterclockwise rotation of $\pi/2$:
$$
\hat{\mathbf e}_{\psi,t}^{(i,j)} \;\coloneq\; R_{\!\pi/2}\,\hat{\mathbf e}_{p,t}^{(i,j)}.
$$
Projecting the relative velocity onto these axes yields the two features: \begin{itemize}
  \item \textit{Radial velocity}:
  $$
  v_{p,t}^{(i,j)} \coloneq \mathbf v_t^{(i,j)}\cdot \hat{\mathbf e}_{p,t}^{(i,j)},
  $$
  \item \textit{Tangential velocity}:
  $$
  v_{\psi,t}^{(i,j)} \coloneq \mathbf v_t^{(i,j)}\cdot \hat{\mathbf e}_{\psi,t}^{(i,j)}.
  $$
\end{itemize}

\end{itemize}
\end{enumerate}

\subsection{Action Space}
At each decision step the agent selects one of three speed advisories: increase, decrease, or hold the current calibrated airspeed. Adjustments are applied in fixed increments of $5$ knots. Since we consider the aircraft to be AAM vehicles with hovering capabilities, the calibrated airspeed is bounded between 0 knots and the aircraft's maximum permitted speed. We denote endpoints of this range as $v_\text{min}=0$ and $v_\text{max}.$

\subsection{Reward Function}\label{sec:reward-function}
To train a policy that issues speed advisories toward aircraft desired speeds while avoiding safety-critical events, we design a reward function that balances positive and negative components. The positive terms reinforce progress toward desirable behavior, whereas the negative terms penalize unsafe or undesirable outcomes. Importantly, the reward is carefully designed to promote safe and adaptive behavior that remains effective in environments beyond those seen during training.

Let $\alpha_{(\cdot)}$ denote a hyperparameter of a reward component. The reward components are then categorized into three groups:  

\begin{enumerate}
    \item \textit{No Conflict:}  
    \begin{enumerate}
        \item \textit{Proximity to desired speed:}  
        If no conflict is expected to occur within a fixed look-ahead horizon $L$, the agent is rewarded for approaching the desired speed:  
        \begin{equation*}
            \mathcal{R}_v(s_{t}, a_{t}) \coloneq \textcolor{ForestGreen}{(+)}\ \alpha_{v} \, \Delta \hat{v}_t,
        \end{equation*}
        where $\Delta\hat{v}_t = 1-\frac{\Delta v_t}{v_{\max}^{} - v_{min}} \in [0,1]$ is the normalized proximity to desired speed.
    \end{enumerate}
    
    \item \textit{In Conflict:}  
    \begin{enumerate}
       \item \textit{Time to intrusion:}  
        If a conflict is expected to occur within a fixed look-ahead horizon $L$, the agent incurs a penalty proportional to the minimum normalized time-to-intrusion:  
        \begin{equation*}
            \mathcal{R}_{\text{conflict}}(s_t, a_t) \coloneq
            \textcolor{red}{(-)}\ \alpha_\text{conflict}\, \hat{T}_{\text{los},t},
        \end{equation*}
        where
        $$
            \hat{T}_{\text{los},t} \coloneq \text{clip}\left(
            \left[\frac{L - \min_{j \in \mathcal{N} \setminus \{i\}} 
            T_{\text{los},t}^{(i,j)}}{L}\right];\ 0, 1\right),
        $$
        and $T_{\text{los},t}^{(i,j)} \in [0, L]$ denotes the extrapolated time to loss of separation between aircraft $i$ and $j$, computed from their relative velocity components.

        \item \textit{Distance to NMAC:}  
        If the agent is already in loss of separation, we impose an additional penalty based on the normalized distance to the nearest intruder:  
        \begin{equation*}
            \mathcal{R}_{\text{los}}(s_t, a_t) \coloneq \textcolor{red}{(-)} \ \alpha_\text{los} \, \hat{d}_t^{(i,j)},
        \end{equation*}
        where the normalized distance between aircraft $i$ and $j$ is defined as
        $$
        \hat{d}^{(i,j)}_t \coloneq \text{clip}\left( \left[ \frac{r_\text{pz} - \min_{j \in \mathcal{N} \setminus \{i\}} \left(d^{(i, j)}_t \right)}{r_\text{pz} - r_\text{nmac}} \right];\ 0, 1\right),
        $$ 
        and $d^{(i,j)}_t$ denotes the Euclidean distance between aircraft $i$ and $j$.
    \end{enumerate}
    \item \textit{In NMAC}:
    If an agent is in a near-mid-air collision, we apply a large penalty to disincentivize this behavior:$$
    \mathcal{R}_\text{nmac} \coloneq \textcolor{red}{(-)}\ \alpha_{nmac}.
    $$ In our setting, we set $\alpha_\text{nmac}$ to $-100.$
\end{enumerate}

By aggregating the individual reward components, the overall reward function takes the compact form:
\begin{equation*}
\small
\mathcal{R}(s_t, a_t) =
\begin{cases}
    \mathcal{R}_v(s_t, a_t), & \text{if \textbf{not in} conflict}, \\[6pt]
    \mathcal{R}_{\text{conflict}}(s_t, a_t) + \mathcal{R}_{\text{los}}(s_t, a_t), & \text{if \textbf{in} conflict,} \\
    \mathcal{R}_\text{nmac}, & \text{if \textbf{in} NMAC.}\\
\end{cases}
\end{equation*}

\vspace{0.1in}

\section{Neural Network Architecture} \label{sec:network-structure}
Figure~\ref{fig:network-architecture} depicts our neural network architecture. It builds on an encoder-transformer design from natural language processing \cite{devlin2019bert}, adapted to operate on intruder tokens rather than word tokens. In this design, ownship information is encoded by a conditioned classifier token that aggregates information from the intruder tokens. We detail each component as follows:

\begin{figure*}[ht]
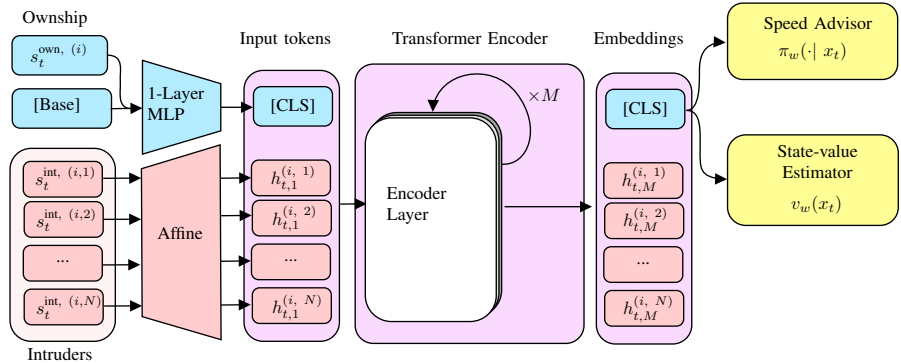
 
    \centering
    \includestandalone[width=0.65\linewidth]{graphics/network-viz}
    \caption{The implemented network adopts an encoder–transformer architecture, where intruder information is represented as intruder tokens. A classifier ($\texttt{[CLS]}$) token, derived from ownship information, is appended to this set. The complete token sequence is processed by a transformer encoder with $M$ layers. The final-layer $\texttt{[CLS]}$ token is used to produce both the policy and the state-value estimates.}
    \label{fig:network-architecture}
\end{figure*}

\underline{Ownship feature adaptation:}
The ownship features are encoded as a conditioned classifier token. To construct this token, we first define a base token as a learnable parameter initialized from a standard Gaussian distribution. This learnable prior is then fused with the ownship information through concatenation, followed by an affine transformation, GELU activation \cite{hendrycks2016gelu}, and layer normalization \cite{ba2016layernorm} to explicitly condition the token on the ego state. The resulting representation serves as the classifier token, denoted as $\texttt{[CLS]}$, which will later serve as an aggregator of information across the token space.

\underline{Intruder feature adaptation:}
The intruder features are encoded as intruder tokens. To obtain these tokens, we individually transform the features of each intruder into the embedding space through an affine transformation with layer normalization. This process produces a set of intruder tokens, one for each intruder.

\underline{Token processing and aggregation:}
After adapting both the ownship and intruder features to the embedding space, we concatenate the $\texttt{[CLS]}$ token with the intruder tokens to form a token set. This set is then processed by an $M$-layer transformer encoder, which applies self-attention \cite{vaswani2017attention}, layer normalization, and GELU activations to produce an attention-refined token set. From the encoder output, we extract the $\texttt{[CLS]}$ token and use it as input to both the policy distribution (speed advisor) and the state-value function estimator heads. 

Here we explicitly use the $\texttt{[CLS]}$ token as a global aggregator rather than pooling over intruder tokens, for three reasons. First, it allows the encoder to gather all intruder information into a single representation. This removes the need for ad hoc pooling schemes over intruder tokens, such as min- or max-pooling. Second, because the $\texttt{[CLS]}$ token is conditioned on ownship features, it carries that context through all encoder layers. Alternative designs would need to inject ownship information later at the transformer heads via concatenation, which may restrict ownship context to the final stages of processing and weaken its influence on earlier, lower-level representations. Third, this design naturally supports a variable number of intruders, including the zero-intruder case, without requiring explicit padding across intruder tokens.


\section{Model Training}\label{sec:training}
This section outlines our training methodology. We begin with a brief overview of the proximal policy optimization algorithm used to train our models, followed by a description of the training environment and then the training details. 

\subsection{Proximal Policy Optimization}
We train our network using the clipped variant of Proximal Policy Optimization (PPO)~\cite{schulman2017proximal}. 
PPO extends the policy gradient framework, where the parameters of a stochastic policy $\pi_w(a \mid s)$ are optimized by ascending the gradient of the expected return. 
In standard policy gradient methods, the update direction is given by
$$
\nabla_w J(w) = \mathbb{E}_t \Big[ \nabla_w \log \pi_w(a_t \mid s_t)\, G_t \Big],
$$
where $G_t$ is the return-to-go and $\mathbb{E}_t$ denotes an expectation over time steps collected under the current policy. This update increases the likelihood of actions with higher returns and decreases it for less beneficial actions.

Large steps in this naive update, however, can destabilize the learning process.
PPO addresses this issue by maximizing a surrogate objective that constrains the deviation between the new and old policies. 
This surrogate objective is defined as:
\begin{equation*}
\max_w \ \hat{\mathbb{E}} \Big[ \min \big( r_t(w)\, \hat{A}_t, \ \text{clip}\big(r_t(w), 1 - \epsilon, 1 + \epsilon\big)\, \hat{A}_t \big) \Big],
\end{equation*}
where
$r_t(w) = \frac{\pi_w(a_t \mid s_t)}{\pi_{w_{\text{old}}}(a_t \mid s_t)}
$
is the probability ratio between the updated and old policies, $\hat{A}_t$ is an estimate of the advantage function (typically computed as the difference between the obtained reward $R_t$ and a learned state-value function $v(s_t)$), and $\epsilon$ is the clipping parameter. The clipping parameter $\epsilon$ prevents overly aggressive updates and leads to more stable training.

\subsection{Training Environment}
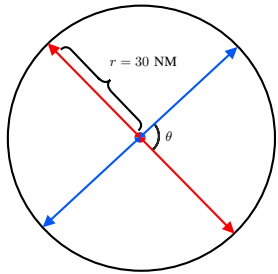
\begin{figure}[h]
    \centering
    \begin{tikzpicture}[x=0.75pt,y=0.75pt,yscale=-1,xscale=1]

\draw  [color={rgb, 255:red, 39; green, 39; blue, 39 }  ,draw opacity=1 ][line width=1.5]  (156.42,135.14) .. controls (156.6,135.33) and (156.77,135.54) .. (156.94,135.74) .. controls (162.93,143.07) and (161.96,153.81) .. (154.81,159.94) -- (154.81,159.94) .. controls (161.96,153.81) and (162.93,143.07) .. (156.94,135.74) .. controls (156.77,135.54) and (156.6,135.33) .. (156.42,135.14) -- cycle ;
\draw  [line width=1.5]  (17.5,148.75) .. controls (17.5,78.75) and (74.25,22) .. (144.25,22) .. controls (214.25,22) and (271,78.75) .. (271,148.75) .. controls (271,218.75) and (214.25,275.5) .. (144.25,275.5) .. controls (74.25,275.5) and (17.5,218.75) .. (17.5,148.75) -- cycle ;
\draw [color={rgb, 255:red, 255; green, 0; blue, 0 }  ,draw opacity=1 ][fill={rgb, 255:red, 0; green, 0; blue, 0 }  ,fill opacity=1 ][line width=1.5]    (58.04,60.74) -- (191.26,197.52) -- (229.96,237.26) ;
\draw [shift={(232.75,240.13)}, rotate = 225.76] [fill={rgb, 255:red, 255; green, 0; blue, 0 }  ,fill opacity=1 ][line width=0.08]  [draw opacity=0] (11.61,-5.58) -- (0,0) -- (11.61,5.58) -- cycle    ;
\draw [shift={(55.25,57.88)}, rotate = 45.76] [fill={rgb, 255:red, 255; green, 0; blue, 0 }  ,fill opacity=1 ][line width=0.08]  [draw opacity=0] (11.61,-5.58) -- (0,0) -- (11.61,5.58) -- cycle    ;
\draw [color={rgb, 255:red, 0; green, 87; blue, 255 }  ,draw opacity=1 ][line width=1.5]    (53.43,231.27) -- (232.57,64.23) ;
\draw [shift={(235.5,61.5)}, rotate = 137] [fill={rgb, 255:red, 0; green, 87; blue, 255 }  ,fill opacity=1 ][line width=0.08]  [draw opacity=0] (11.61,-5.58) -- (0,0) -- (11.61,5.58) -- cycle    ;
\draw [shift={(50.5,234)}, rotate = 317] [fill={rgb, 255:red, 0; green, 87; blue, 255 }  ,fill opacity=1 ][line width=0.08]  [draw opacity=0] (11.61,-5.58) -- (0,0) -- (11.61,5.58) -- cycle    ;
\draw  [line width=1.5]  (141.43,141.36) .. controls (144.8,138.14) and (144.88,134.84) .. (141.66,131.46) -- (116.54,105.12) .. controls (111.93,100.3) and (111.32,96.28) .. (114.7,93.05) .. controls (111.32,96.28) and (107.33,95.48) .. (102.73,90.65)(104.8,92.82) -- (77.6,64.31) .. controls (74.38,60.94) and (71.08,60.86) .. (67.71,64.08) ;
\draw  [draw opacity=0][fill={rgb, 255:red, 255; green, 0; blue, 0 }  ,fill opacity=1 ] (137.69,147.99) .. controls (137.69,145.21) and (140.06,142.96) .. (142.99,142.96) -- (142.99,147.99) -- cycle ; \draw  [draw opacity=0] (137.69,147.99) .. controls (137.69,145.21) and (140.06,142.96) .. (142.99,142.96) ;  
\draw  [draw opacity=0][fill={rgb, 255:red, 0; green, 77; blue, 255 }  ,fill opacity=1 ] (142.99,153.02) .. controls (142.99,153.02) and (142.99,153.02) .. (142.99,153.02) .. controls (140.06,153.02) and (137.69,150.77) .. (137.69,147.99) -- (142.99,147.99) -- cycle ; \draw  [draw opacity=0] (142.99,153.02) .. controls (142.99,153.02) and (142.99,153.02) .. (142.99,153.02) .. controls (140.06,153.02) and (137.69,150.77) .. (137.69,147.99) ;  
\draw  [draw opacity=0][fill={rgb, 255:red, 0; green, 87; blue, 255 }  ,fill opacity=1 ] (142.91,142.96) .. controls (142.91,142.96) and (142.91,142.96) .. (142.91,142.96) .. controls (142.91,142.96) and (142.91,142.96) .. (142.91,142.96) .. controls (145.83,142.96) and (148.21,145.21) .. (148.21,147.99) .. controls (148.21,147.99) and (148.21,147.99) .. (148.21,147.99) -- (142.91,147.99) -- cycle ; \draw  [draw opacity=0] (142.91,142.96) .. controls (142.91,142.96) and (142.91,142.96) .. (142.91,142.96) .. controls (142.91,142.96) and (142.91,142.96) .. (142.91,142.96) .. controls (145.83,142.96) and (148.21,145.21) .. (148.21,147.99) .. controls (148.21,147.99) and (148.21,147.99) .. (148.21,147.99) ;  
\draw  [draw opacity=0][fill={rgb, 255:red, 255; green, 0; blue, 0 }  ,fill opacity=1 ] (148.21,147.99) .. controls (148.21,150.77) and (145.83,153.02) .. (142.91,153.02) -- (142.91,147.99) -- cycle ; \draw  [draw opacity=0] (148.21,147.99) .. controls (148.21,150.77) and (145.83,153.02) .. (142.91,153.02) ;

\draw (165.97,140.65) node [anchor=north west][inner sep=0.75pt]    {$\theta $};
\draw (113,69.9) node [anchor=north west][inner sep=0.75pt]    {$r=30\ \text{NM}$};

\end{tikzpicture}
   \caption{The simulated training environment consists of two routes whose endpoints are uniformly sampled within a circle of radius $30$ nautical miles. The routes may intersect at angles $\theta \in [0^\circ, 180^\circ]$ and can be traversed in either direction.}
    \label{fig:training-environment}
\end{figure}

We train our policy in a simulated air traffic control environment built upon the BlueSky simulator \cite{hoekstra2016bluesky}. The environment models a circular flight sector in which two routes intersect at a central point. Each route extends over a fixed distance with a sector radius of approximately $30$ nautical miles. Agents act at one-second intervals. To improve generalization, we randomize the intersection angle between the routes and enforce a minimum separation of $5$ nautical miles between the route endpoints. A visualization of the training environment is provided in Figure~\ref{fig:training-environment}.

Unlike traditional fixed-wing settings, we consider aircraft with VTOL capabilities, which have both forward motion and the ability to hover. This additional degree of freedom not only modifies the state dynamics but also introduces the challenge of hovering behavior, which may lead to deadlock (i.e., the environment may not terminate).

To ensure that the environment terminates, we consider two conditions under which aircraft may be removed from the environment. The first occurs when an aircraft takes too long to reach its destination. For this, we precompute an expected time of arrival for each aircraft based on its desired speed, plus a 20-minute buffer, which we set due to battery constraints. If an aircraft exceeds this precomputed time, it is removed from the environment. The second scenario arises in the event of a near mid-air collision. In such a case, the infringing aircraft are removed and a substantial penalty (see Section~\ref{sec:reward-function}) is imposed.

\subsection{Training Details}
We train the network described in Section~\ref{sec:network-structure} using PPO across eight parallel environments, collectively simulating approximately $75$ days of traffic. A fixed-length rollout buffer aggregates trajectories from all agents, with each environment generating sequences of $4096$ seconds. Transitions are randomly shuffled and sampled to optimize the PPO objective. To stabilize training, we apply value clipping and advantage normalization. 

In total, $200$ updates are performed on a transformer-encoder network with an embedding dimension of $128$ and an encoder feed-forward dimension of $512$. We further train three model variants with encoder depths of $1$, $2$, and $3$ layers. To ensure fair evaluation and reduce sensitivity to initialization, each model is trained three times with different random seeds. A summary of the relevant neural network configurations and hyperparameters is provided in Table~\ref{tab:network-structure} and Table~\ref{tab:hyperparameters}, respectively.

\begin{table}[t]
  \centering
  \caption{The transformer-based neural network configurations employed in our experiments.}
  \label{tab:network-structure}
  \setlength{\tabcolsep}{8pt}
  \begin{tabular}{@{}l l@{}}
    \toprule
    Component & Setting \\
    \midrule
    Encoder layers               & \([1,\,2,\,3]\) \\
    Token/embedding dimension    & 128 \\
    Feed-forward (FFN) dimension & 512 \\
    Attention heads              & 16 \\
    \bottomrule
  \end{tabular}
\end{table}
\begin{table}[h]
    \centering
    \caption{Hyperparameters employed in the PPO training of the transformer network.}
    \begin{tabular}{l p{0.13\linewidth}}
        \toprule
        Hyperparameter & Value \\
        \midrule
        Number of updates & $200$ \\
        Parallel environments & $8$ \\
        Time horizon ($T$) & $4096$ \\
        Batch size & $128$ \\
        Update epochs & $4$ \\
        Discount factor ($\gamma$) & $0.99$ \\
        GAE-$\lambda$ & $0.95$ \\
        Clip coefficient ($\epsilon$) & $0.2$ \\
        Entropy coefficient & $0.01$ \\
        Value function coefficient & $0.5$ \\
        Maximum gradient norm & $0.5$ \\
        Advantage normalization & $\checkmark$ \\
        Value function clipping & $\checkmark$ \\
        \bottomrule
    \end{tabular}
    \label{tab:hyperparameters}
\end{table}

\section{Simulation Experiments}\label{sec:results-discussions}
This section presents the results of our simulation experiments. In this section, we will first examine the training curves to compare the three neural network configurations, each with 1, 2, and 3 encoder layers. After that, we will assess each model's ability to adapt to unseen scenarios in both structured and unstructured airspaces. Lastly, we will compare our transformer-based architecture with a neural network architecture constructed solely with attention mechanisms.

\begin{figure}[t]
    \centering
    \begin{subfigure}{0.3\linewidth}
        \centering
        \tikzset{every picture/.style={line width=0.75pt}} 

\begin{tikzpicture}[x=0.75pt,y=0.75pt,yscale=-1,xscale=1]

\draw  [line width=2.25]  (288.5,143.25) .. controls (288.5,73.25) and (345.25,16.5) .. (415.25,16.5) .. controls (485.25,16.5) and (542,73.25) .. (542,143.25) .. controls (542,213.25) and (485.25,270) .. (415.25,270) .. controls (345.25,270) and (288.5,213.25) .. (288.5,143.25) -- cycle ;
\draw [color={rgb, 255:red, 255; green, 0; blue, 0 }  ,draw opacity=1 ][line width=2.25]    (326.25,52.38) -- (415.25,143.25) ;
\draw [color={rgb, 255:red, 255; green, 0; blue, 0 }  ,draw opacity=1 ][line width=2.25]    (323.33,231) -- (415.25,143.25) ;
\draw [shift={(415.25,143.25)}, rotate = 316.33] [color={rgb, 255:red, 255; green, 0; blue, 0 }  ,draw opacity=1 ][fill={rgb, 255:red, 255; green, 0; blue, 0 }  ,fill opacity=1 ][line width=2.25]      (0, 0) circle [x radius= 5.36, y radius= 5.36]   ;
\draw [color={rgb, 255:red, 255; green, 0; blue, 0 }  ,draw opacity=1 ][line width=2.25]    (288.5,143.25) -- (542,143.25) ;
\draw  [color={rgb, 255:red, 255; green, 79; blue, 79 }  ,draw opacity=1 ][fill={rgb, 255:red, 255; green, 79; blue, 79 }  ,fill opacity=1 ][line width=2.25]  (352.56,143.31) -- (346.19,147.19) -- (346.19,139.44) -- cycle ;
\draw  [color={rgb, 255:red, 255; green, 79; blue, 79 }  ,draw opacity=1 ][fill={rgb, 255:red, 255; green, 79; blue, 79 }  ,fill opacity=1 ][line width=2.25]  (367.53,94.2) -- (360.46,91.81) -- (366.4,86.83) -- cycle ;
\draw  [color={rgb, 255:red, 255; green, 79; blue, 79 }  ,draw opacity=1 ][fill={rgb, 255:red, 255; green, 79; blue, 79 }  ,fill opacity=1 ][line width=2.25]  (369.29,187.13) -- (367.52,194.37) -- (362.04,188.89) -- cycle ;
\draw  [color={rgb, 255:red, 255; green, 79; blue, 79 }  ,draw opacity=1 ][fill={rgb, 255:red, 255; green, 79; blue, 79 }  ,fill opacity=1 ][line width=2.25]  (483.06,143.31) -- (476.69,147.19) -- (476.69,139.44) -- cycle ;
\draw [color={rgb, 255:red, 0; green, 0; blue, 0 }  ,draw opacity=1 ][line width=2.25]    (344.23,22.68) .. controls (390.81,-5.08) and (438.06,-5.3) .. (484.99,22.04) ;
\draw [shift={(488.6,24.2)}, rotate = 211.5] [fill={rgb, 255:red, 0; green, 0; blue, 0 }  ,fill opacity=1 ][line width=0.08]  [draw opacity=0] (14.29,-6.86) -- (0,0) -- (14.29,6.86) -- cycle    ;
\draw [shift={(339.8,25.4)}, rotate = 327.62] [fill={rgb, 255:red, 0; green, 0; blue, 0 }  ,fill opacity=1 ][line width=0.08]  [draw opacity=0] (14.29,-6.86) -- (0,0) -- (14.29,6.86) -- cycle    ;

\end{tikzpicture}
        \caption{}
        \label{subfig:case1}
    \end{subfigure}
    \hfill
    \begin{subfigure}{0.3\linewidth}
        \centering
        \tikzset{every picture/.style={line width=0.75pt}} 

\begin{tikzpicture}[x=0.75pt,y=0.75pt,yscale=-1,xscale=1]

\draw [color={rgb, 255:red, 255; green, 0; blue, 0 }  ,draw opacity=1 ][line width=2.25]    (308.5,161.25) -- (562,161.25) ;
\draw [color={rgb, 255:red, 0; green, 87; blue, 255 }  ,draw opacity=1 ][line width=2.25]    (498.6,272) -- (498.6,50.8) ;
\draw [shift={(498.6,161.4)}, rotate = 270] [color={rgb, 255:red, 0; green, 87; blue, 255 }  ,draw opacity=1 ][fill={rgb, 255:red, 0; green, 87; blue, 255 }  ,fill opacity=1 ][line width=2.25]      (0, 0) circle [x radius= 5.36, y radius= 5.36]   ;
\draw  [line width=2.25]  (308.5,161.25) .. controls (308.5,91.25) and (365.25,34.5) .. (435.25,34.5) .. controls (505.25,34.5) and (562,91.25) .. (562,161.25) .. controls (562,231.25) and (505.25,288) .. (435.25,288) .. controls (365.25,288) and (308.5,231.25) .. (308.5,161.25) -- cycle ;
\draw [color={rgb, 255:red, 255; green, 0; blue, 0 }  ,draw opacity=1 ][line width=2.25]    (346.25,70.38) -- (435.25,161.25) ;
\draw [color={rgb, 255:red, 255; green, 0; blue, 0 }  ,draw opacity=1 ][line width=2.25]    (343.33,249) -- (435.25,161.25) ;
\draw [shift={(435.25,161.25)}, rotate = 316.33] [color={rgb, 255:red, 255; green, 0; blue, 0 }  ,draw opacity=1 ][fill={rgb, 255:red, 255; green, 0; blue, 0 }  ,fill opacity=1 ][line width=2.25]      (0, 0) circle [x radius= 5.36, y radius= 5.36]   ;
\draw  [color={rgb, 255:red, 255; green, 79; blue, 79 }  ,draw opacity=1 ][fill={rgb, 255:red, 255; green, 79; blue, 79 }  ,fill opacity=1 ][line width=2.25]  (372.56,161.31) -- (366.19,165.19) -- (366.19,157.44) -- cycle ;
\draw  [color={rgb, 255:red, 255; green, 79; blue, 79 }  ,draw opacity=1 ][fill={rgb, 255:red, 255; green, 79; blue, 79 }  ,fill opacity=1 ][line width=2.25]  (387.53,112.2) -- (380.46,109.81) -- (386.4,104.83) -- cycle ;
\draw  [color={rgb, 255:red, 255; green, 79; blue, 79 }  ,draw opacity=1 ][fill={rgb, 255:red, 255; green, 79; blue, 79 }  ,fill opacity=1 ][line width=2.25]  (389.29,205.13) -- (387.52,212.37) -- (382.04,206.89) -- cycle ;
\draw  [color={rgb, 255:red, 255; green, 79; blue, 79 }  ,draw opacity=1 ][fill={rgb, 255:red, 255; green, 79; blue, 79 }  ,fill opacity=1 ][line width=2.25]  (473.86,161.31) -- (467.49,165.19) -- (467.49,157.44) -- cycle ;
\draw  [color={rgb, 255:red, 0; green, 87; blue, 255 }  ,draw opacity=1 ][fill={rgb, 255:red, 0; green, 87; blue, 255 }  ,fill opacity=1 ][line width=2.25]  (498.48,207.73) -- (502.34,214.11) -- (494.59,214.09) -- cycle ;
\draw  [color={rgb, 255:red, 0; green, 87; blue, 255 }  ,draw opacity=1 ][fill={rgb, 255:red, 0; green, 87; blue, 255 }  ,fill opacity=1 ][line width=2.25]  (498.88,102.93) -- (502.74,109.31) -- (494.99,109.29) -- cycle ;
\draw  [color={rgb, 255:red, 255; green, 79; blue, 79 }  ,draw opacity=1 ][fill={rgb, 255:red, 255; green, 79; blue, 79 }  ,fill opacity=1 ][line width=2.25]  (538.66,161.31) -- (532.29,165.19) -- (532.29,157.44) -- cycle ;
\draw [color={rgb, 255:red, 0; green, 0; blue, 0 }  ,draw opacity=1 ][line width=2.25]    (364.23,42.68) .. controls (410.81,14.92) and (458.06,14.7) .. (504.99,42.04) ;
\draw [shift={(508.6,44.2)}, rotate = 211.5] [fill={rgb, 255:red, 0; green, 0; blue, 0 }  ,fill opacity=1 ][line width=0.08]  [draw opacity=0] (14.29,-6.86) -- (0,0) -- (14.29,6.86) -- cycle    ;
\draw [shift={(359.8,45.4)}, rotate = 327.62] [fill={rgb, 255:red, 0; green, 0; blue, 0 }  ,fill opacity=1 ][line width=0.08]  [draw opacity=0] (14.29,-6.86) -- (0,0) -- (14.29,6.86) -- cycle    ;

\end{tikzpicture}
        \caption{}
        \label{subfig:case2}
    \end{subfigure}
    \hfill
    \begin{subfigure}{0.3\linewidth}
        \centering
        \tikzset{every picture/.style={line width=0.75pt}} 

\begin{tikzpicture}[x=0.75pt,y=0.75pt,yscale=-1,xscale=1]

\draw  [line width=2.25]  (297.5,138.25) .. controls (297.5,68.25) and (354.25,11.5) .. (424.25,11.5) .. controls (494.25,11.5) and (551,68.25) .. (551,138.25) .. controls (551,208.25) and (494.25,265) .. (424.25,265) .. controls (354.25,265) and (297.5,208.25) .. (297.5,138.25) -- cycle ;
\draw [color={rgb, 255:red, 255; green, 34; blue, 34 }  ,draw opacity=1 ][line width=2.25]    (396.4,89.4) -- (369.6,251.8) ;
\draw [shift={(396.4,89.4)}, rotate = 99.37] [color={rgb, 255:red, 255; green, 34; blue, 34 }  ,draw opacity=1 ][fill={rgb, 255:red, 255; green, 34; blue, 34 }  ,fill opacity=1 ][line width=2.25]      (0, 0) circle [x radius= 5.36, y radius= 5.36]   ;
\draw [color={rgb, 255:red, 0; green, 87; blue, 255 }  ,draw opacity=1 ][line width=2.25]    (301.6,105.4) -- (414.8,180.2) ;
\draw [shift={(414.8,180.2)}, rotate = 33.46] [color={rgb, 255:red, 0; green, 87; blue, 255 }  ,draw opacity=1 ][fill={rgb, 255:red, 0; green, 87; blue, 255 }  ,fill opacity=1 ][line width=2.25]      (0, 0) circle [x radius= 5.36, y radius= 5.36]   ;
\draw [color={rgb, 255:red, 0; green, 87; blue, 255 }  ,draw opacity=1 ][line width=2.25]    (506,41) -- (414.8,180.2) ;
\draw  [color={rgb, 255:red, 255; green, 79; blue, 79 }  ,draw opacity=1 ][fill={rgb, 255:red, 255; green, 79; blue, 79 }  ,fill opacity=1 ][line width=2.25]  (389.22,133.74) -- (386.22,126.91) -- (393.9,127.93) -- cycle ;
\draw [color={rgb, 255:red, 255; green, 34; blue, 34 }  ,draw opacity=1 ][line width=2.25]    (396.4,89.4) -- (549.6,159.8) ;
\draw  [color={rgb, 255:red, 255; green, 79; blue, 79 }  ,draw opacity=1 ][fill={rgb, 255:red, 255; green, 79; blue, 79 }  ,fill opacity=1 ][line width=2.25]  (491.58,133.17) -- (499.01,132.51) -- (495.59,139.46) -- cycle ;
\draw  [color={rgb, 255:red, 0; green, 87; blue, 255 }  ,draw opacity=1 ][fill={rgb, 255:red, 0; green, 87; blue, 255 }  ,fill opacity=1 ][line width=2.25]  (348.14,136.13) -- (340.75,135.09) -- (345.67,129.09) -- cycle ;
\draw  [color={rgb, 255:red, 0; green, 87; blue, 255 }  ,draw opacity=1 ][fill={rgb, 255:red, 0; green, 87; blue, 255 }  ,fill opacity=1 ][line width=2.25]  (437.11,146.18) -- (437.17,153.64) -- (430.51,149.66) -- cycle ;
\draw  [color={rgb, 255:red, 255; green, 79; blue, 79 }  ,draw opacity=1 ][fill={rgb, 255:red, 255; green, 79; blue, 79 }  ,fill opacity=1 ][line width=2.25]  (378.42,198.54) -- (375.42,191.71) -- (383.1,192.73) -- cycle ;
\draw  [color={rgb, 255:red, 0; green, 87; blue, 255 }  ,draw opacity=1 ][fill={rgb, 255:red, 0; green, 87; blue, 255 }  ,fill opacity=1 ][line width=2.25]  (480,80.23) -- (479.81,87.69) -- (473.29,83.49) -- cycle ;

\end{tikzpicture}
        \caption{}
        \label{subfig:case3}
    \end{subfigure}

    \caption{Evaluation of the model under three scenarios simulating structured and unstructured airspace. 
    Case~(\subref{subfig:case1}) and Case~(\subref{subfig:case2}) represent structured airspace, but with rotational variability. Case~(\subref{subfig:case3}) serves as a proxy for unstructured airspace, which includes one intermediary waypoint. For visualization purposes, we only show two routes in Case~(\subref{subfig:case3}). However, the possible number of routes is dictated by the spawning scheme defined in Table~\ref{tab:spawning-details}.} \label{fig:evaluation-scenarios}
\end{figure}
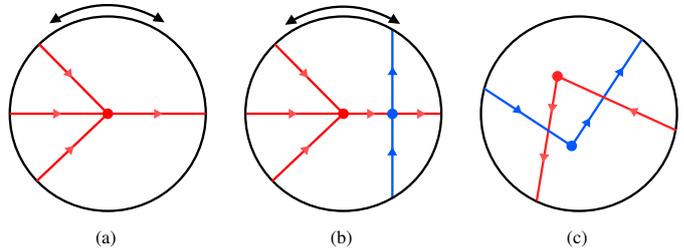

\subsection{Training Curves}
We trained three transformer configurations with 1, 2, and 3 encoder layers for 200 updates, running three independent runs per configuration.  Throughout training, we evaluated model performance using two metrics:
\begin{enumerate}
    \item \textit{$\lambda$-return}: the bootstrapped return as calculated by the Generalized Advantage Estimation (GAE) \cite{schulman2018gae}; and
    \item \textit{Policy entropy}: a measure of the stochasticity (i.e., predictability) of the learned policies.
\end{enumerate}

\begin{figure*}[ht]
    \centering
    \begin{subfigure}{0.49\linewidth}
        \centering
        \includegraphics[width=\linewidth]{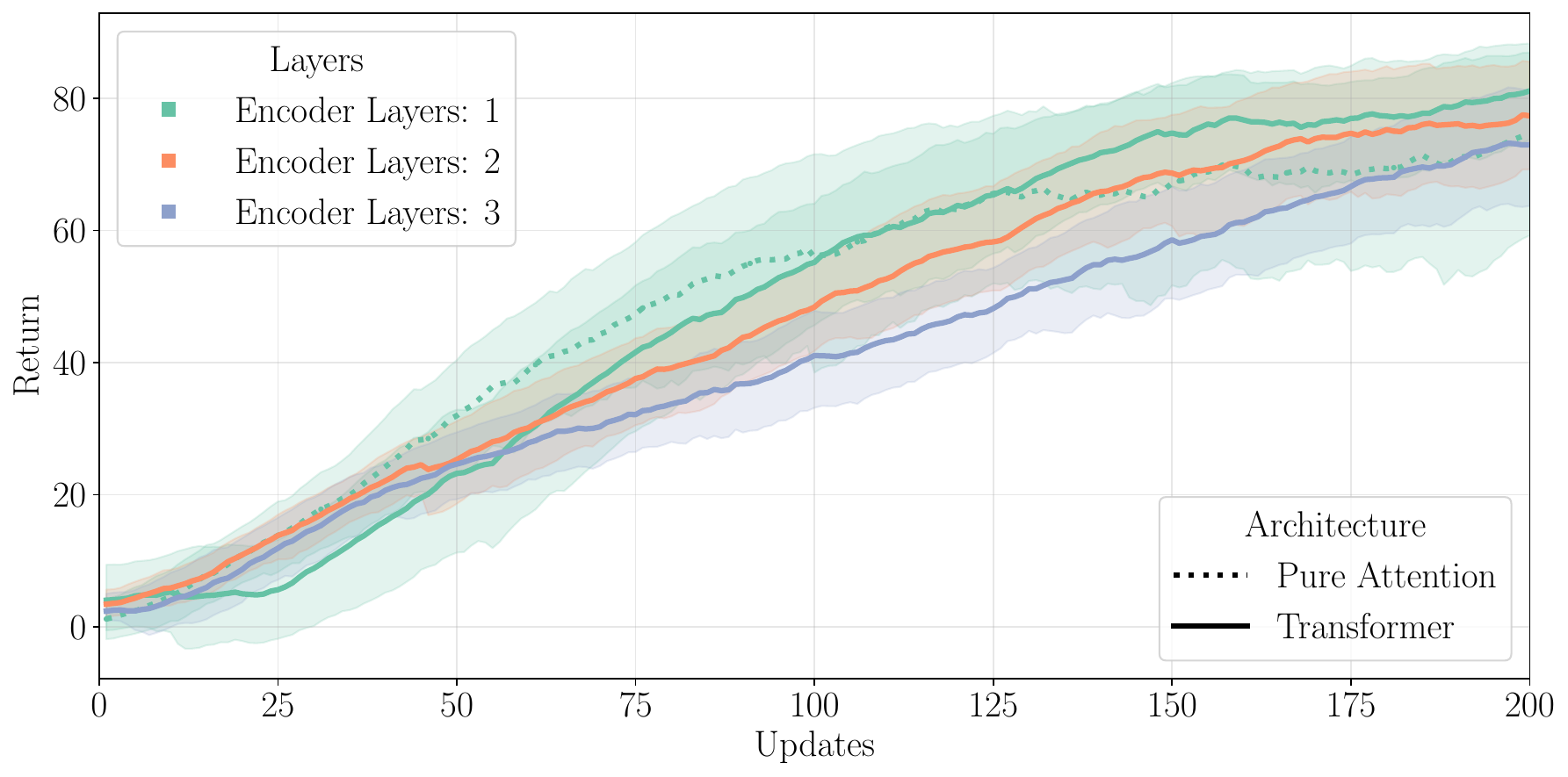}
        \caption{}
        \label{fig:training-return}
    \end{subfigure}\hfill
    \begin{subfigure}{0.49\linewidth}
        \centering\includegraphics[width=\linewidth]{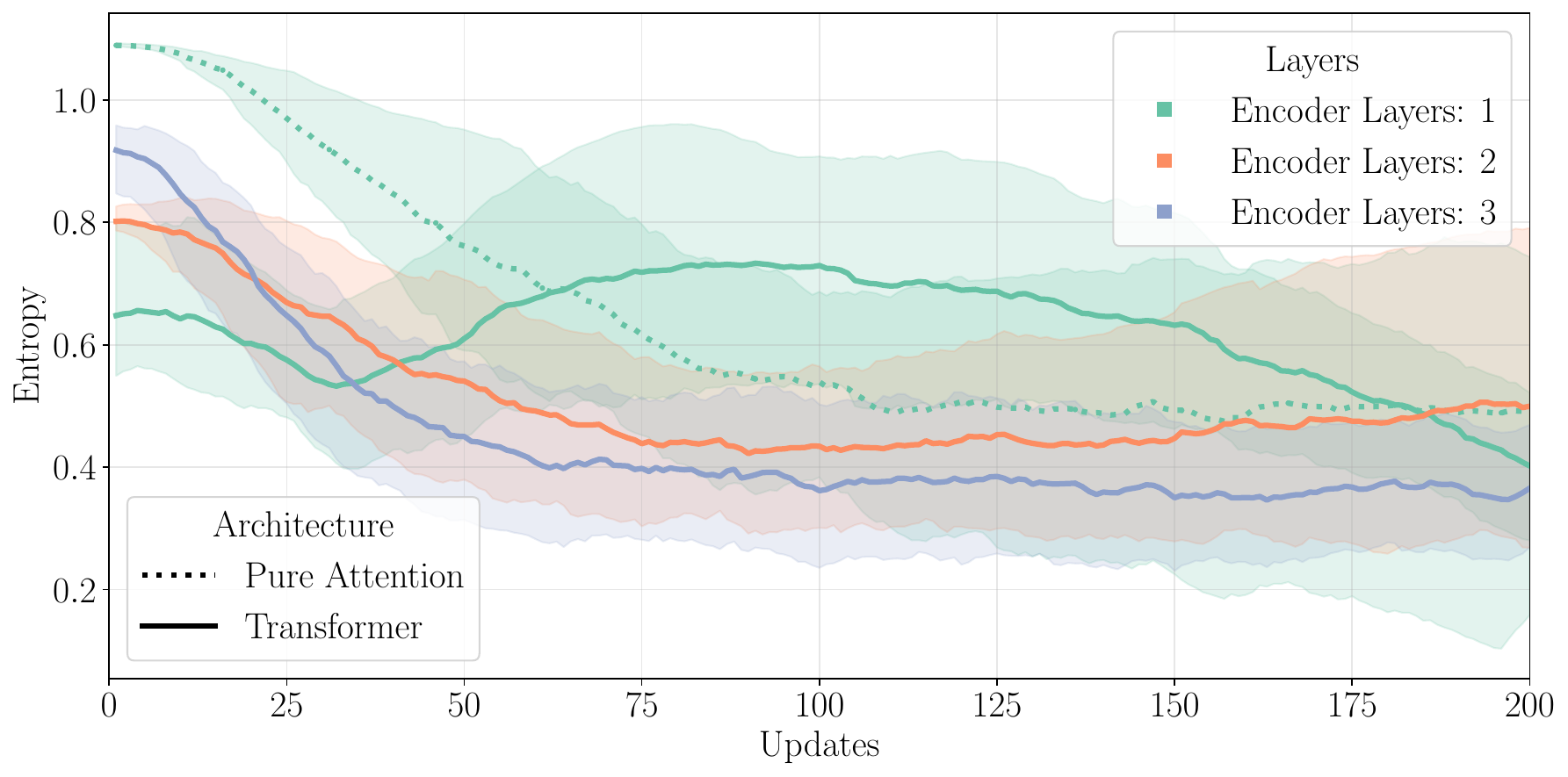}
        \caption{}
        \label{fig:training-entropy}
    \end{subfigure}
    \caption{The average $\lambda$-returns (\subref{fig:training-return}) and entropy (\subref{fig:training-entropy}) of the transformer network during training, shown for configurations with $1$, $2$, and $3$ encoder layers. Each configuration was trained using three random seeds. Solid lines represent the mean values, while shaded regions indicate the range between the minimum and maximum values. Lines are smoothed with an exponential moving average with $\alpha=0.05$.}
    \label{fig:training-metrics}
\end{figure*}

We begin with Figure~\ref{fig:training-return}, which plots the bootstrapped returns over the $200$ training updates. In this plot, we observe that all encoder configurations show a steady rise in $\lambda$-returns over time. The curves start near $0$ and rise to about $75$ to $83$ across configurations. Among them, the 1-layer encoder attains the highest final $\lambda$-return at $83.1$, followed by the 2-layer at $78.8$, and the 3-layer at $75.1$.

The entropy of the policies is shown in Figure~\ref{fig:training-entropy}. Across all transformer network configurations, entropy decreases over time, reflecting convergence to lower-entropy policies. The $1$-layer encoder exhibits a marked deviation between roughly the $25^\text{th}$ and $160^\text{th}$ updates, where entropy rises and stabilizes near $0.75$ before returning to a lower value. The $2$-layer encoder shows the broadest range across runs, with a high upper bound of about $0.8$ that emerges toward the end of training. In contrast, the $3$-layer encoder displays a steadier decline with tighter variability. By the final updates, all models converge to comparable entropy values, ranging from around $0.35$ to $0.5$.

\subsection{Unseen Sector Structure Configurations}
After training the three different transformer configurations on the training environment (see Figure~\ref{fig:training-environment}), we then evaluated the trained model's performance on three unseen airspace structures, as visualized in Figure~\ref{fig:evaluation-scenarios}: 
\begin{itemize}
    \item Case~(\subref{subfig:case1}): a single merge point where three routes converge; 
    \item Case~(\subref{subfig:case2}): a single merge point of three routes with a downstream four-way intersection; and
    \item Case~(\subref{subfig:case3}): an airspace with procedurally generated routes whose endpoints are sampled on the sector boundary with one intermediate waypoint placed randomly within the sector. 
\end{itemize}  
The first two cases represent structured airspace, whereas the third serves as a proxy for unstructured airspace. To assess the generalizability of the model and to ensure rotational invariance, Case~(\subref{subfig:case1}) and  Case~(\subref{subfig:case2}) are randomly rotated by an angle uniformly sampled from $0^\circ$ to $360^\circ$. Because Case~(\subref{subfig:case3}) includes randomly sampled intermediate waypoints, some instances may be infeasible without requiring all aircraft to hover. To resolve this issue, we allow the early termination of an aircraft’s trajectory if it is in a loss-of-separation and can decelerate to a stop. Additional details regarding the sampling distributions of the aircraft are provided in Table~\ref{tab:spawning-details}.

\begin{table}[h!]
    \centering
    \caption{Aircraft distributions in each of the environments.}    \label{tab:spawning-details}

    \small
     \begin{tabular}{l | c c c}
    \toprule
    Environment & $N$ aircraft & Spacing (s) & Desired speed (kts) \\
    \midrule
    Training & $U(1, 20)$   & $U(60, 1200)$ & $U(60, 120)$ \\
    Case~(\subref{subfig:case1}) & $U(1, 20)$   & $U(60, 1200)$ & $110$ \\
    Case~(\subref{subfig:case2}) & $U(1, 20)$   & $U(60, 1200)$ & $110$ \\
    Case~(\subref{subfig:case3}) & $U(1, 10)$   & $U(60, 1200)$ & $U(60, 120)$ \\
    \bottomrule
\end{tabular}

    \parbox{\linewidth}{\footnotesize
        \footnotesize
        $\dagger$ Here, $N$ aircraft denotes the number of aircraft spawned in the flight sector. $U(a,b)$ denotes a draw from the uniform distribution on $[a,b]$. 
    }
\end{table}

\subsection{Performance on Unseen Sector Structures}
To analyze model performance in unseen sector configurations, we focus on three metrics: \textit{(i)} the number of near mid-air collisions (NMACs), \textit{(ii)} the cumulative time in loss of separation (LOS) across all aircraft pairs, summed over all timesteps, and \textit{(iii)} adherence to desired speeds, measured as the proportion of timesteps in which aircraft are within $10$ knots of their desired cruising speed. Benchmark results are summarized in Table~\ref{tab:model-safety}, averaged across 1{,}000 evaluation episodes with three independently trained sets of weights for each neural network configuration.

\begin{table*}[ht]
    \caption{Performance and safety metrics of the different network configurations across 1000 episodes.}
    \label{tab:model-safety}
    \centering
    \definecolor{pastelorange}{rgb}{1.0, 0.7, 0.28}
\definecolor{mossgreen}{rgb}{0.68, 0.87, 0.68}

{\large
\renewcommand{\arraystretch}{1.1}
\begin{tabular}{cc|ccc|ccc|ccc|ccc}
    \toprule
    \multicolumn{2}{c|}{} & \multicolumn{3}{c|}{Pure Attention} & \multicolumn{3}{c|}{\textbf{1-Layer  Transformer}} & \multicolumn{3}{c|}{2-Layer  Transformer} & \multicolumn{3}{c}{3-Layer  Transformer} \\
    Environment & Density & NMACs & Speed Adher. & LoS Time & NMACs & Speed Adher. & LoS Time & NMACs & Speed Adher. & LoS Time & NMACs & Speed Adher. & LoS Time \\
    \midrule

    \multirow{4}{*}{Training} & $\leq$5 & $0.008 \pm 0.125$ & $0.653 \pm 0.175$ & $108.338 \pm 527.907$ & $0.000 \pm 0.000$ & $0.843 \pm 0.182$ & $119.300 \pm 570.666$ & $0.000 \pm 0.000$ & $0.819 \pm 0.170$ & $190.182 \pm 813.784$ & $0.011 \pm 0.145$ & $0.868 \pm 0.132$ & $92.371 \pm 461.763$ \\
     & 6-10 & $0.080 \pm 0.437$ & $0.578 \pm 0.205$ & $1771.115 \pm 4060.318$ & $0.008 \pm 0.124$ & $0.709 \pm 0.212$ & $1119.789 \pm 2463.339$ & $0.002 \pm 0.067$ & $0.646 \pm 0.235$ & $2215.736 \pm 4111.380$ & $0.049 \pm 0.329$ & $0.726 \pm 0.202$ & $1017.803 \pm 2379.765$ \\
     & 11+ & $0.337 \pm 0.927$ & $0.371 \pm 0.218$ & $11727.380 \pm 14699.918$ & $0.000 \pm 0.000$ & $0.490 \pm 0.206$ & $4059.592 \pm 5150.277$ & $0.000 \pm 0.000$ & $0.418 \pm 0.217$ & $6455.796 \pm 6595.136$ & $0.091 \pm 0.417$ & $0.539 \pm 0.218$ & $3161.449 \pm 4144.970$ \\
     & Average & $0.081 \pm 0.453$ & $0.583 \pm 0.211$ & $2191.441 \pm 6422.559$ & $0.005 \pm 0.097$ & $0.736 \pm 0.223$ & $1016.157 \pm 2585.519$ & $0.001 \pm 0.052$ & $0.685 \pm 0.241$ & $1874.085 \pm 4020.136$ & $0.040 \pm 0.294$ & $0.756 \pm 0.206$ & $898.693 \pm 2344.195$ \\
    \midrule
    \multirow{4}{*}{Case~(\subref{subfig:case1})} & $\leq$5 & $0.000 \pm 0.000$ & $0.671 \pm 0.110$ & $3.524 \pm 43.188$ & $0.000 \pm 0.000$ & $0.828 \pm 0.081$ & $0.515 \pm 5.325$ & $0.000 \pm 0.000$ & $0.789 \pm 0.128$ & $3.973 \pm 19.792$ & $0.000 \pm 0.000$ & $0.858 \pm 0.082$ & $131.431 \pm 804.446$ \\
     & 6-10 & $0.001 \pm 0.045$ & $0.645 \pm 0.138$ & $55.615 \pm 326.128$ & $0.000 \pm 0.000$ & $0.759 \pm 0.075$ & $13.360 \pm 145.403$ & $0.001 \pm 0.045$ & $0.696 \pm 0.152$ & $97.871 \pm 650.170$ & $0.010 \pm 0.143$ & $0.781 \pm 0.071$ & $387.216 \pm 1531.111$ \\
     & 11+ & $0.014 \pm 0.166$ & $0.560 \pm 0.169$ & $650.686 \pm 1550.735$ & $0.002 \pm 0.070$ & $0.572 \pm 0.111$ & $78.825 \pm 192.645$ & $0.000 \pm 0.000$ & $0.542 \pm 0.175$ & $744.886 \pm 2412.845$ & $0.063 \pm 0.363$ & $0.660 \pm 0.075$ & $1544.828 \pm 3230.180$ \\
     & Average & $0.005 \pm 0.097$ & $0.622 \pm 0.152$ & $225.269 \pm 916.817$ & $0.001 \pm 0.037$ & $0.712 \pm 0.123$ & $30.327 \pm 158.063$ & $0.001 \pm 0.037$ & $0.659 \pm 0.175$ & $273.182 \pm 1412.005$ & $0.025 \pm 0.230$ & $0.750 \pm 0.095$ & $713.279 \pm 2217.769$ \\
    \midrule
    \multirow{4}{*}{Case~(\subref{subfig:case2})} & $\leq$5 & $0.000 \pm 0.000$ & $0.677 \pm 0.114$ & $0.000 \pm 0.000$ & $0.000 \pm 0.000$ & $0.839 \pm 0.066$ & $0.000 \pm 0.000$ & $0.000 \pm 0.000$ & $0.814 \pm 0.094$ & $3.889 \pm 18.900$ & $0.000 \pm 0.000$ & $0.868 \pm 0.047$ & $4.791 \pm 25.383$ \\
     & 6-10 & $0.002 \pm 0.066$ & $0.593 \pm 0.148$ & $156.961 \pm 509.802$ & $0.000 \pm 0.000$ & $0.700 \pm 0.080$ & $37.113 \pm 113.309$ & $0.000 \pm 0.000$ & $0.641 \pm 0.129$ & $206.468 \pm 863.310$ & $0.019 \pm 0.194$ & $0.731 \pm 0.082$ & $249.403 \pm 895.354$ \\
     & 11+ & $0.023 \pm 0.216$ & $0.496 \pm 0.159$ & $1245.955 \pm 1747.369$ & $0.000 \pm 0.000$ & $0.551 \pm 0.079$ & $215.158 \pm 408.550$ & $0.000 \pm 0.000$ & $0.482 \pm 0.125$ & $962.233 \pm 2023.816$ & $0.177 \pm 0.594$ & $0.599 \pm 0.081$ & $1271.186 \pm 2338.114$ \\
     & Average & $0.010 \pm 0.141$ & $0.558 \pm 0.159$ & $555.713 \pm 1250.620$ & $0.000 \pm 0.000$ & $0.652 \pm 0.110$ & $96.785 \pm 268.506$ & $0.000 \pm 0.000$ & $0.590 \pm 0.151$ & $462.197 \pm 1418.890$ & $0.073 \pm 0.390$ & $0.688 \pm 0.105$ & $598.839 \pm 1626.276$ \\
    \midrule
    \multirow{4}{*}{Case~(\subref{subfig:case3})} & $\leq$3 & $0.000 \pm 0.000$ & $0.535 \pm 0.210$ & $149.869 \pm 527.145$ & $0.000 \pm 0.000$ & $0.771 \pm 0.179$ & $223.657 \pm 679.895$ & $0.000 \pm 0.000$ & $0.759 \pm 0.150$ & $306.918 \pm 802.779$ & $0.000 \pm 0.000$ & $0.774 \pm 0.153$ & $217.779 \pm 659.005$ \\
     & 4-6 & $0.014 \pm 0.166$ & $0.537 \pm 0.179$ & $1415.154 \pm 1793.011$ & $0.003 \pm 0.078$ & $0.688 \pm 0.109$ & $1480.423 \pm 1883.388$ & $0.000 \pm 0.000$ & $0.667 \pm 0.104$ & $1996.577 \pm 2150.875$ & $0.009 \pm 0.135$ & $0.678 \pm 0.108$ & $1602.484 \pm 2009.602$ \\
     & 7+ & $0.018 \pm 0.191$ & $0.522 \pm 0.179$ & $3525.313 \pm 3123.919$ & $0.009 \pm 0.132$ & $0.614 \pm 0.093$ & $3626.186 \pm 2983.981$ & $0.000 \pm 0.000$ & $0.600 \pm 0.085$ & $4420.622 \pm 3195.953$ & $0.018 \pm 0.186$ & $0.617 \pm 0.083$ & $3681.700 \pm 3049.316$ \\
     & Average & $0.012 \pm 0.154$ & $0.534 \pm 0.185$ & $1485.924 \pm 2133.126$ & $0.003 \pm 0.082$ & $0.692 \pm 0.132$ & $1569.346 \pm 2183.789$ & $0.000 \pm 0.000$ & $0.676 \pm 0.122$ & $2002.420 \pm 2447.275$ & $0.009 \pm 0.131$ & $0.687 \pm 0.125$ & $1653.816 \pm 2273.653$ \\
    \bottomrule
    \toprule
    Overall & Average & $0.027 \pm 0.256$ & \cellcolor{pastelorange}$0.574 \pm 0.181$ & $1114.587 \pm 3556.437$ & $0.002 \pm 0.066$ & $0.698 \pm 0.156$ & \cellcolor{mossgreen} $678.154 \pm 1817.575$ & \cellcolor{mossgreen}$0.001 \pm 0.032$ & $0.652 \pm 0.181$ & \cellcolor{pastelorange}$1152.971 \pm 2676.009$ & \cellcolor{pastelorange}$0.037 \pm 0.279$ & \cellcolor{mossgreen}$0.720 \pm 0.143$ & $966.157 \pm 2173.696$ \\
    \bottomrule
    
    \label{tab:environment-details}
\end{tabular}
}

    \parbox{\linewidth}{\footnotesize
        \footnotesize
        $\dagger$ Here, density refers to the maximum number of aircraft appearing in the sector at any point in an episode, while speed adherence refers to the proportion of time in which aircraft are within 10 knots of their desired speed. Cells highlighted in green indicate the best overall performance, while cells highlighted in orange indicate the worst overall performance. The configuration in bold indicates the best model.
    }
\end{table*}

We begin by analyzing the model's capability to prevent NMACs. As shown in Table~\ref{tab:model-safety}, the different transformer configurations indicate relatively safe separation capabilities. In particular, our $1$ and $2$ encoder layer networks show near-zero NMACs, each with average incursion values of $0.002$ and $0.001$, respectively. By contrast, the three-layer network incurs a slightly higher occurrence, with an average number of NMACs of $0.037$.  

Continuing to LoS, we observe that all transformer configurations generally incur fewer LoS events in the structured airspace of Case~(\subref{subfig:case1}) and Case~(\subref{subfig:case2}), while showing noticeably higher incursions in the unstructured airspace of Case~(\subref{subfig:case3}). For the $1$-layer encoder, the overall mean LoS value is $678.154$, with substantially lower averages of $30.327$ in Case~(\subref{subfig:case1}) and $96.785$ in Case~(\subref{subfig:case2}). The $2$-layer encoder reaches an overall mean LoS of $1152.971$, driven upward by the high average of $2002.420$ in Case~(\subref{subfig:case3}). The $3$-layer encoder averages $966.157$ overall and performs worst in the structured settings of Case~(\subref{subfig:case1}) and Case~(\subref{subfig:case2}). We note that in all scenarios the aircraft are restricted to speed adjustments only, which likely limits their ability to mitigate safety incursions, particularly in Case~(\subref{subfig:case3}).

Lastly, in terms of adherence to the desired cruising speeds, the three-layer encoder attains the closest desired speed adherence with a proportion of $72\%$. Meanwhile, the one and two-layer encoders have a slightly lower value of $69.8\%$ and $65.2\%$, respectively. This reduced proportion may be attributed to the higher safety factors (lower NMACs and LoS), which prompt the aircraft to decelerate or accelerate more often to prevent incursions from occurring.

\subsection{Comparisons to Pure Attention}
We compare our approach with a pure attention model \cite{brittain2022marl}. As summarized in Table~\ref{tab:model-safety}, our transformer-based neural network, particularly the single-layer encoder configuration, achieves a higher level of safety with fewer NMACs and generally shorter time in LoS. In addition, all our transformer network configurations with 1, 2, and 3 encoder layers tracked the desired cruising speed more closely than the pure attention baseline.

\section{Discussions} \label{sec:discussions}
We study safe-separation assurance with two objectives: (\emph{i}) maintaining safe separation and (\emph{ii}) minimizing deviation from the desired cruising speed. We cast the problem in a multi-agent reinforcement learning setting and introduce a reformulated state space and neural network architecture. To improve generalization, we trained the models on varied airspace structures with different intersection angles and traffic densities. Our neural network architecture is a transformer network containing a $\texttt{[CLS]}$ token conditioned on ownship features. Furthermore, we benchmarked models with $1$, $2$, and $3$ encoder layer configurations and compared them with an attention-only baseline on three unseen airspace structures.

Our training curves indicate that increasing the number of encoder layers does not reliably improve model performance. The single-layer encoder achieved the highest returns, followed by the two-layer encoder configuration, with the three-layer configuration obtaining the lowest returns. This pattern suggests that a single encoder layer may be sufficient to learn safe separation policies with speed-only adjustments. 

Two factors may explain the weaker performance of the deeper models. First, the small number of training updates may have disadvantaged the deeper models, which have a greater number of parameters. These deeper models may require a greater number of updates to match the performance of the shallower networks. Second, speed-only actions may lead deeper models to over-optimize the dense speed reward. In contrast, smaller models appear less prone to exploiting these rewards and instead learn to avoid the more consequential conflict penalties. Taken together, these factors are consistent with the higher returns we observed for the single-layer network.

A similar trend is observed when considering the safety metrics.  Across $1,000$ evaluation episodes over three flight sectors, the single-layer transformer network achieved the best combined safety performance with a near-zero count of near mid-air collisions (NMACs) and the lowest average time in loss of separation (LoS) among the tested configurations. It was most reliable in the structured airspaces of Case~(\subref{subfig:case1}) and Case~(\subref{subfig:case2}), where it also maintained close adherence to desired speed. However, performance declined slightly in the unstructured airspace of Case~(\subref{subfig:case3}), which is consistent with the limitations of a speed-only action space for separation assurance. The two-layer encoder configurations yielded similar safety outcomes but slightly longer LoS times. The three-layer configuration performed the worst, with the highest occurrence of NMACs.

Finally, we showed that our transformer architecture outperformed an attention-only baseline. Specifically, the $1$-layer encoder configuration achieved lower NMAC counts, lower LoS time, and closer adherence to desired speeds than the baseline. This result may stem from the additional use of multi-head attention and the use of a $\texttt{[CLS]}$ token conditioned on ownship information. However, this hypothesis may require further investigation.

\section{Conclusion}\label{sec:conclusion}
This study demonstrates that a multi-agent reinforcement learning framework, supported by a carefully designed relative state representation and a tailored reward function, can resolve aircraft conflicts while maintaining aircraft close to their desired cruising speeds. Our evaluation of encoder depths revealed that a transformer architecture with a $1$-layer encoder outperformed the deeper configurations, achieving a near-zero NMAC rate and shorter time in loss of separation. Additionally, we demonstrated that this configuration outperforms a baseline model constructed solely with pure attention. Future work could enhance the effectiveness of our methodology through refined training strategies, such as increasing the number of intersecting routes, allowing for heading or altitude changes, or modifying the reward function. Overall, our approach represents a promising step toward enhanced MARL formulation and neural network architectures for aircraft separation assurance in future high-density AAM operations.

\section*{Acknowledgment}

This work was supported by the National Science Foundation under Grant No. $2312092$. Any opinions, findings, and conclusions or recommendations expressed in this material are those of the authors and do not necessarily reflect the views of the National Science Foundation.

%
\printbibliography

@article{brittain2022marl,
  author   = {Brittain, Marc and Wei, Peng},
  title    = {Scalable Autonomous Separation Assurance with Heterogeneous Multi-Agent Reinforcement Learning},
  journal  = {IEEE Transactions on Automation Science and Engineering},
  year     = {2022},
  volume   = {19},
  number   = {4},
  pages    = {2837--2848}
}

@inproceedings{hoekstra2016bluesky,
  title={{BlueSky ATC} simulator project: an open data and open source approach},
  author={Hoekstra, Jacco M and Ellerbroek, Joost},
  booktitle={Proceedings of the 7th international conference on research in air transportation},
  volume={131},
  pages={132},
  year={2016},
  organization={FAA/Eurocontrol Washington, DC, USA}
}

@inproceedings{groot2022using,
  title     = {Using Relative State Transformer Models for Multi-Agent Reinforcement Learning in Air Traffic Control},
  author    = {Groot, D. J. and Ellerbroek, Joost and Hoekstra, J. M.},
  booktitle = {SESAR Innovation Days},
  year      = {2022},
  month     = nov,
  address   = {Seville, Spain}
}

@article{schulman2017proximal,
  title={Proximal policy optimization algorithms},
  author={Schulman, John and Wolski, Filip and Dhariwal, Prafulla and Radford, Alec and Klimov, Oleg},
  journal={arXiv preprint arXiv:1707.06347},
  year={2017}
}

@inproceedings{devlin2019bert,
    title = "{BERT}: Pre-training of Deep Bidirectional Transformers for Language Understanding",
    author = "Devlin, Jacob  and
      Chang, Ming-Wei  and
      Lee, Kenton  and
      Toutanova, Kristina",
    booktitle = "Proceedings of the 2019 Conference of the North {A}merican Chapter of the Association for Computational Linguistics: Human Language Technologies, Volume 1 (Long and Short Papers)",
    month = jun,
    year = "2019",
    address = "Minneapolis, Minnesota",
    publisher = "Association for Computational Linguistics",
    pages = "4171--4186"
}

@article{vaswani2017attention,
  title={Attention is all you need},
  author={Vaswani, Ashish and Shazeer, Noam and Parmar, Niki and Uszkoreit, Jakob and Jones, Llion and Gomez, Aidan N and Kaiser, {\L}ukasz and Polosukhin, Illia},
  journal={{Advances in neural information processing systems}},
  volume={30},
  year={2017}
}

@article{hendrycks2016gelu,
  title={Gaussian error linear units ({GELUs})},
  author={Hendrycks, Dan and Gimpel, Kevin},
  journal={arXiv preprint arXiv:1606.08415},
  year={2016}
}

@article{ba2016layernorm,
  title={Layer normalization},
  author={Ba, Jimmy Lei and Kiros, Jamie Ryan and Hinton, Geoffrey E},
  journal={arXiv preprint arXiv:1607.06450},
  year={2016}
}

@article{groot2025comparing,
  title={Comparing attention-based methods with long short-term memory for state encoding in reinforcement learning-based separation management},
  author={Groot, DJ and Ellerbroek, J and Hoekstra, JM},
  journal={Engineering Applications of Artificial Intelligence},
  volume={159},
  pages={111592},
  year={2025},
  publisher={Elsevier}
}

@inproceedings{schulman2018gae,
  title     = {High-Dimensional Continuous Control Using Generalized Advantage Estimation},
  author    = {Schulman, John and Moritz, Philipp and Levine, Sergey and Jordan, Michael I. and Abbeel, Pieter},
  booktitle = {Proceedings of the International Conference on Learning Representations (ICLR)},
  year      = {2016}
}

@inproceedings{Pham2019Machine,
  author    = {Duc-Thinh Pham and Ngoc Phu Tran and Sameer Alam and Vu Duong and Daniel Delahaye},
  title     = {A Machine Learning Approach for Conflict Resolution in Dense Traffic Scenarios with Uncertainties},
  booktitle = {Thirteenth USA/Europe Air Traffic Management Research and Development Seminar},
  year      = {2019},
  publisher = {},            

}

@article{hochreiter1997long,
  title={Long short-term memory},
  author={Hochreiter, Sepp and Schmidhuber, J{\"u}rgen},
  journal={Neural computation},
  volume={9},
  number={8},
  pages={1735--1780},
  year={1997},
  publisher={MIT press}
}

@inproceedings{cho-2014-learning,
    title = "Learning Phrase Representations using {RNN} Encoder{--}Decoder for Statistical Machine Translation",
    author = {Cho, Kyunghyun  and
      van Merri{\"e}nboer, Bart  and
      Gulcehre, Caglar  and
      Bahdanau, Dzmitry  and
      Bougares, Fethi  and
      Schwenk, Holger  and
      Bengio, Yoshua},
    editor = "Moschitti, Alessandro  and
      Pang, Bo  and
      Daelemans, Walter",
    booktitle = "Proceedings of the 2014 Conference on Empirical Methods in Natural Language Processing ({EMNLP})",
    month = oct,
    year = "2014",
    address = "Doha, Qatar",
    publisher = "Association for Computational Linguistics",
    pages = "1724--1734"
}

@inproceedings{reimers2019sentence,
  title     = {Sentence-{BERT}: Sentence Embeddings using Siamese {BERT}-Networks},
  author    = {Reimers, Nils and Gurevych, Iryna},
  booktitle = {Proceedings of the 2019 Conference on Empirical Methods in Natural Language Processing
               and the 9th International Joint Conference on Natural Language Processing (EMNLP-IJCNLP)},
  year      = {2019}
}

@article{dosovitskiy2020image,
  title={An image is worth 16x16 words: Transformers for image recognition at scale},
  author={Dosovitskiy, Alexey},
  journal={arXiv preprint arXiv:2010.11929},
  year={2020}
}

@INPROCEEDINGS{brittain2019marl,
  author={Brittain, Marc and Wei, Peng},
  booktitle={2019 IEEE Intelligent Transportation Systems Conference (ITSC)}, 
  title={Autonomous Separation Assurance in An High-Density En Route Sector: A Deep Multi-Agent Reinforcement Learning Approach}, 
  year={2019},
  volume={},
  number={},
  pages={3256-3262}}


\end{document}